\documentclass{article}

\PassOptionsToPackage{numbers, compress}{natbib}

\usepackage[preprint]{preprint}

\usepackage[utf8]{inputenc}
\usepackage[T1]{fontenc}
\usepackage{hyperref}
\usepackage{url}
\usepackage{booktabs}
\usepackage{amsfonts}
\usepackage{amsmath}
\usepackage{amssymb}
\usepackage{nicefrac}
\usepackage{microtype}
\usepackage{xcolor}
\usepackage{graphicx}
\usepackage{makecell}

\title{Mechanistic Interpretability of \\ EEG Foundation Models via Sparse Autoencoders}

\author{%
  William Lehn-Schi\o ler\textsuperscript{1,2,3} \And
  Magnus Ruud Kj\ae r\textsuperscript{2} \And
  Rahul Thapa\textsuperscript{4, 5} \And
  Magnus Guldberg Pedersen\textsuperscript{1,3} \And
  Anton Mosquera Storgaard\textsuperscript{1,3} \And
  Nick Williams\textsuperscript{6} \And
 Radu Gătej
 \textsuperscript{1} \And
  Tue Lehn-Schi\o ler\textsuperscript{1} \And
  Andreas Brink-Kjær\textsuperscript{2}
   \And
  Sadasivan Puthusserypady\textsuperscript{2} \And
  S\'andor Beniczky\textsuperscript{7,8} \And
  James Zou \textsuperscript{4,5}\And
  Lars Kai Hansen\textsuperscript{3} 
}

\begin{document}

\maketitle

\vspace{-2em}

\begin{center}
\small
\textsuperscript{1}BrainCapture, Kongens Lyngby, Denmark \\
\textsuperscript{2}DTU Health Tech, Technical University of Denmark, Kongens Lyngby, Denmark \\
\textsuperscript{3}DTU Compute, Technical University of Denmark, Kongens Lyngby, Denmark \\
\textsuperscript{4}Department of Biomedical Data Science, Stanford University, Stanford, CA, USA \\
\textsuperscript{5}Department of Computer Science, Stanford University, Stanford, CA, USA \\
\textsuperscript{6}Seer Medical, Melbourne, Australia \\
\textsuperscript{7}Filadelfia Epilepsy Hospital, Dianalund, Denmark \\
\textsuperscript{8}University Hospital of Copenhagen, Copenhagen, Denmark
\end{center}
\vspace{0.5em}

\begin{abstract}
EEG foundation models achieve state-of-the-art clinical performance, yet the internal computations driving their predictions remain opaque: a barrier to clinical trust. We apply TopK Sparse Autoencoders (SAEs) across three architecturally distinct EEG transformers: SleepFM, REVE, and LaBraM to extract sparse feature dictionaries from their embeddings. By grounding these features in a clinical taxonomy (abnormality, age, sex, and medication), we benchmark monosemanticity and entanglement across architectures. A single hyperparameter procedure, driven by an intrinsic dictionary health audit, transfers robustly across all three architectures. Via concept steering, we introduce a "target vs. off-target" probe area metric to quantify steering selectivity and reveal three operational regimes: selectively steerable, encoded but entangled, and non-encoded. This framework exposes critical representational failures: "wrecking-ball" interventions that collapse global model performance, and clinical entanglements, such as age–pathology confounding, where it is impossible to suppress one concept without corrupting the other. Finally, a spectral decoder maps these interventions back to the amplitude spectrum, translating latent manipulations into physiologically interpretable frequency signatures, such as pathological slow-wave suppression and $\alpha$-band restoration.
\end{abstract}

\section{Introduction}
\label{sec:intro}

EEG foundation models pre-trained on large corpora achieve strong performance across a range of tasks: sleep staging, pathology detection, and brain-computer interface decoding~\cite{thapa2024sleepfm,elouahidi2025reve, jiang2024labram, kostas2021bendr, lehnschioler2026pretraining}. Yet the internal computations that produce these predictions remain opaque. Unlike image models, where attention maps already give a coarse visual story, EEG models operate on multichannel time-series whose clinically relevant events: spindles, K-complexes, $\delta$ waves, epileptiform discharges, are transient, channel-distributed, and defined by decades of human expert knowledge~\cite{beniczky2017score}. Understanding \emph{what} an EEG foundation model has learned, and \emph{where} it has learned it, is both a scientific question and a clinical-trust requirement.

Mechanistic interpretability \cite{elhage2021mathematical,elhage2022superposition} addresses this by reverse-engineering the functional roles of individual directions inside a network. The central obstacle is \textit{superposition}~\cite{elhage2022superposition}: a transformer layer with $d$ dimensions can represent up to $N \gg d$ features by co-activating them sparsely, making individual neurons uninterpretable. Sparse Autoencoders (SAEs) \cite{bricken2023monosemanticity, templeton2024scaling, lieberum2024gemma} have emerged as the dominant tool for recovering these features in language models. A small but growing literature ports the toolkit to medical
imaging~\cite{renzulli2025medsae, nakka2025mammosae}, protein
language models~\cite{simon2025interplm}, and transformers
trained on neural signals~\cite{freeman2025beyond,
kalnare2025mechanistic}. Despite this progress, EEG foundation models remain
unaddressed; no prior work has bridged the full path from a
frozen EEG encoder to sparse feature dictionaries, clinical
concept attribution, and spectrum-level mechanistic explanation
across architectures.

\paragraph{Contributions.}
We make the following contributions:
\begin{enumerate}
\item \textbf{A Cross-Architecture Interpretability Pipeline:} We introduce a unified framework for interpreting EEG transformers that bridges the gap between high-dimensional embeddings and clinical physiology. The pipeline integrates layer-wise TopK SAEs, concept attribution via Testing with Concept Activation Vectors (TCAV), and concept steering, utilizing a single hyperparameter-selection procedure that remains robust across three distinct architectures: SleepFM, REVE, and LaBraM. 
\\

The interactive framework is available at 

\begin{center}
\url{https://ai.braincapture.dk/mechanistic-interpretability/}
\end{center}

\item \textbf{The Clinical Semanticity Taxonomy:} We propose a taxonomy to audit encoder representations by partitioning clinical features into three regimes: Separable (monosemantic), Entangled (polysemantic co-activations), and Dead (semantically uninformative/inactive). This serves as a tool to identify latent clinical biases where labels like age or medication co-activate with pathology.

\item \textbf{Selectivity via Concept Steering:} We formalize a probe-based selectivity metric; calculating the area between target and off-target "steering curves" to evaluate the fidelity of model interventions. This allows us to distinguish between selective concept removal and "wrecking-ball" interventions, where feature suppression inadvertently collapses the entire embedding space.

\item \textbf{Mechanistic Spectral Explanations:} By re-decoding steered activations through a spectral decoder, we transform abstract latent embedding into human interpretable space. This provides mechanistic evidence for model attributions, allowing experts to verify that a pathology prediction is driven by physiologically relevant features.
\end{enumerate}

\textbf{The Code} is available and openly accessible at \url{https://github.com/BrainCapture/mechanistic-interpretability-for-eeg-foundation-models/tree/preprint}.

\section{Background}

\subsection{EEG Foundation Models}
\label{sec:eeg_models}

We study three architecturally distinct EEG transformer models spanning two different self-supervised pretraining objectives (multi-modal contrastive learning and masked-token prediction); these are summarized in Table~\ref{tab:encoders}. All three encoders are subsequently finetuned end-to-end on the same binary normal/abnormal classification target which is later used as a clinical concept. Performance metrics can be found in Table~\ref{tab:layer-sweep-summary}.

\subsection{TopK Sparse Autoencoders}
\label{sec:background_sae}

A TopK SAE~\cite{makhzani2014ksparse} trained on activations $\mathbf{a} \in \mathbb{R}^d$ learns
\begin{equation}
\mathbf{z} = \mathrm{TopK}\left(\mathbf{W}_{\text{enc}}\,\frac{\mathbf{a} - \boldsymbol{\mu_\ell}}{\boldsymbol{\sigma_\ell}}, k\right),
\quad
\hat{\mathbf{a}} = \mathbf{W}_{\text{dec}}\,\mathbf{z} + \mathbf{b}_{\text{dec}},
\label{eq:sae}
\end{equation}
where $\mathbf{z} \in \mathbb{R}^N$ and $\mathbf{W}_{\text{dec}} \in \mathbb{R}^{d \times N}$ has unit-norm columns (decoder directions $\mathbf{w}_i \in \mathbb{R}^d$); $N \triangleq d \cdot E$ and $E$ is the expansion rate. The conventional SAE formulation uses an $L_1$ sparsity penalty~\cite{cunningham2023sparse}, which encourages but does not guarantee sparsity. We instead use the TopK hard constraint~\cite{templeton2024scaling}: pre-activations are computed for all $N$ features (after standardizing the input using $\mu_\ell$ and $\sigma_\ell$, the per-dimension mean and standard deviation computed over the training-split activations at layer $\ell$), then all but the $k$ largest are set to zero. Exactly $k$ features fire per token by construction, making sparsity directly controllable.

\begin{table}[h]
  \caption{\textbf{Encoder overview.} Three architecturally distinct EEG transformers spanning different SSL objectives: multi-modal contrastive learning (SleepFM), masked-token reconstruction (REVE), and masked spectrum prediction (LaBraM) are inspected. SleepFM is pretrained on PSG data (EEG, ECG, EMG, and respiratory) with a multi-modal contrastive objective, while REVE and LaBraM are pretrained on EEG alone. All three are subsequently finetuned end-to-end on the same binary normal/abnormal target.}
  \label{tab:encoders}
  \centering
  \small
  \resizebox{\textwidth}{!}{
\begin{tabular}{lllccclr}
    \toprule
    \textbf{Encoder} & \textbf{Architecture} & \textbf{$d$} & \textbf{Layers} & \textbf{Token length} & \textbf{SSL objective} & \textbf{Modality} & \textbf{$\approx$Hours} \\
    \midrule
    SleepFM~\cite{thapa2024sleepfm} & SetTransformer & 128 & 3 & 1 sec & MMC (InfoNCE)~\cite{oord2018cpc} & PSG & 585{,}000 \\
    REVE~\cite{elouahidi2025reve}   & Transformer    & 512 & 22 & 1 sec & MAE~\cite{he2022mae} & EEG & 60{,}000 \\
    LaBraM~\cite{jiang2024labram}   & BERT-style     & 200 & 12 & 1 sec & MSP (VQ / BEiT)~\cite{bao2022beit} & EEG & 2{,}500 \\
    \bottomrule
\end{tabular}
  }
\end{table}

\subsection{TCAV and Concept Attribution}
\label{sec:background_tcav}

Concept Activation Vectors~\cite{kim2018tcav} define a concept $C$ via a
linear classifier separating $X_C$ from $X_{\neg C}$ in representation
space, yielding a unit-norm direction $\mathbf{v}_C \in \mathbb{R}^d$.
We additionally train a linear probe $\mathbf{w}$ on SAE activations
$\mathbf{z} \in \mathbb{R}^{N}$, assigning each example a sensitivity
$S_C(\mathbf{x}) = \mathbf{w}^\top \mathbf{z}(\mathbf{x})$.
The TCAV score is the fraction of concept examples with positive
sensitivity:
\begin{equation}
    \mathrm{TCAV}_C =
    \frac{|\{\mathbf{x} \in X_C : S_C(\mathbf{x}) > 0\}|}{|X_C|},
\label{eq:tcav}
\end{equation}

with significance assessed against $N_{\mathrm{rand}}=50$ random-label
null CAVs.


\section{Methodology}

\begin{figure}[h]
  \centering
  \includegraphics[width=\linewidth]{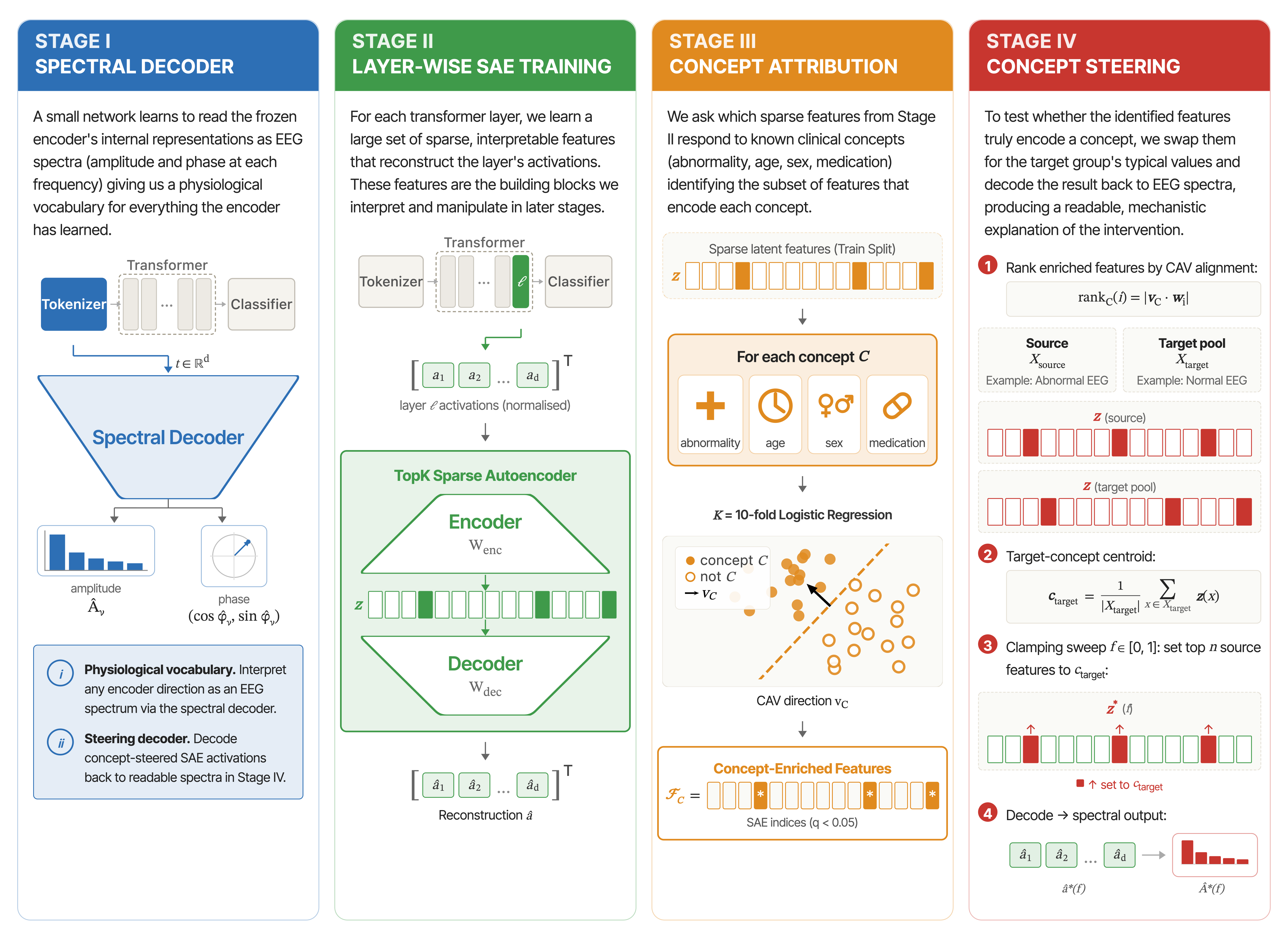}
  \caption{\textbf{Pipeline overview.} Starting from a frozen EEG foundation model: (Stage I) A shallow MLP \emph{spectral decoder} translates token embeddings back into a human interpretable space. (Stage II) For each transformer layer, a TopK SAE recovers a sparse, over-complete feature dictionary from normalized encoder activations. (Stage III) SAE features are mapped to known clinical concepts using TCAV. (Stage IV) \emph{Concept steering} substitutes the top-$n$ concept-enriched feature activations with the target-group centroid, decoding the intervention through the SAE and spectral decoder to produce a mechanistically grounded, spectrum-level explanation.}
  \label{fig:pipeline}
\end{figure}

The pipeline (Figure~\ref{fig:pipeline}) runs in four stages on a frozen encoder. Stages I--III use established interpretability components (linear probing~\cite{alain2017understanding, belinkov2022probing}, dictionary learning~\cite{cunningham2023sparse}, concept attribution~\cite{Madsen_2023}); the novel contribution is Stage IV (concept steering), which uses the dictionary and the spectral decoder to produce mechanistic, spectrum-level explanations of any TCAV attribution.  

\subsection{Dataset and Preprocessing}
\label{sec:data}

We use a clinical 27-lead EEG dataset collected using BrainCapture's BC-1 comprising $3{,}036$ subjects \cite{enkhtsetseg2026kenya}. The cohort is $57.8\%$ male and includes both pediatric and adult populations. Recordings averaged $33.8$ minutes, with $30.2\%$ ($917$ sessions) identified as abnormal; of these, $76.4\%$ showed epileptiform discharges. EEG signals are preprocessed using an evolved version of~\cite{gjolbye2024speed}. The pipeline applies band-pass (0.5-70Hz) and notch (50Hz) filtering before resampling to the encoder's native frequency: $128$~Hz for SleepFM and $200$~Hz for REVE and LaBraM. Recordings are segmented into non-overlapping 60-second windows. Subjects are
partitioned into 80\% train/10\% validation/10\% test sets via scikit-learn's \texttt{GroupShuffleSplit} with random seed 42.

\subsection{Stage I: Spectral Decoder}
\label{sec:stage-i}

The spectral decoder (SD) is a shallow MLP that maps a frozen encoder token embedding $\mathbf{t} \in \mathbb{R}^d$ to amplitude and phase for each frequency bin:
\begin{equation}
\hat{A}_\nu,\;(\cos \hat{\phi}_\nu,\, \sin \hat{\phi}_\nu) = \mathrm{SD}(\mathbf{t}),
\quad \nu = 1, \ldots, F.
\label{eq:sd}
\end{equation}
$F = 64$ bins, $0$--$64$\,Hz, for all encoders (consistent with the
$70$\,Hz low-pass filter applied in preprocessing; Section \ref{sec:data}). Phase is parameterised as a unit-vector pair to eliminate the $2\pi$ discontinuity. SleepFM tokens are channel-averaged; REVE and LaBraM tokens are per-channel-per-second.

The spectral decoder serves two roles: (i) it provides a physiological vocabulary for interpreting any direction $\mathbf{w} \in \mathbb{R}^d$ via $\mathrm{SD}(\mathbf{w})$, and (ii) it is the final stage of concept steering (Section~\ref{sec:steering}).

\subsection{Stage II: Layer-wise SAE Training}
\label{sec:stage-ii}

For each encoder layer $\ell$ we collect all available training-split token activations, normalise per-dimension, and train a TopK SAE with $E \cdot d$ features, where $d$ is the encoder embedding dimension and $E$ is the \emph{expansion ratio} controlling dictionary size relative to the encoder. We scale the sparsity budget with the dictionary size, $k = k_0 \cdot E$ with $k_0 = 8$, so that the per-token active fraction $k / N = k_0 / d$ is constant per encoder across the expansion sweep. This couples the two hyperparameters into a single $E$-axis and ensures that comparisons across $E$ are at matched relative sparsity rather than at matched absolute $k$. The reconstruction objective is MSE on normalised activations; dead neurons (never firing in 500 steps) are periodically resampled following~\cite{templeton2024scaling}. 


\subsection{Stage III: Concept Attribution via TCAV}
\label{sec:stage-iii}

For each concept $C \in \{\textit{abnormality, age group, sex, medication
(psychiatric), medication (ASM)}\}$, we obtain $\mathbf{v}_C$ via
$K_{\mathrm{fold}}=10$-fold $L_2$-regularised logistic regression on
dense layer-$\ell$ activations $\mathbf{a}$, then fit probe $\mathbf{w}$
on $\mathbf{z}$ and compute $\mathrm{TCAV}_C$ per Eq.~\ref{eq:tcav}.
For feature enrichment, we apply a one-sided $z$-test with
Benjamini--Hochberg correction ($q < 0.05$, $N$ simultaneous hypotheses)
to identify concept-enriched SAE features, which form the ranking pool
for Stage~IV.                                                    
\subsection{Stage IV: Concept Steering}\label{sec:steering}

Concept steering provides interventional evidence for the representational fidelity of the SAE dictionary, evaluated against a model-agnostic \emph{selectivity} criterion. Given a target clinical concept $C$ (e.g., age) and a secondary off-target concept (e.g., abnormality), the protocol includes:

\begin{enumerate}
    \item \textbf{Rank by CAV alignment.} The concept direction is mapped onto the sparse dictionary by projecting it onto each SAE decoder direction $\mathbf{w}_i \in \mathbb{R}^d$, giving                         the alignment-based ranking
    \begin{equation}
         \mathrm{rank}_C(i) = |\mathbf{v}_C \cdot \mathbf{w}_i|
    \label{eq:rank_features}
    \end{equation}
    
    \item \textbf{Compute the target-concept centroid.} We define the centroid $\mathbf{c}_{\text{target}} \in \mathbb{R}^N$ as the empirical mean of the SAE feature activations across the target dataset $X_{\text{target}}$ (e.g., the mean of the ``normal'' pool when steering abnormal $\to$ normal):
    \begin{equation}
        \mathbf{c}_{\text{target}} = \frac{1}{|X_{\text{target}}|} \sum_{\mathbf{x} \in X_{\text{target}}} \mathbf{z}(\mathbf{x})
    \label{eq:c_target}
    \end{equation}
    
    \item \textbf{Clamping features.} For a given intervention fraction $f \in [0, 1]$ and a source sample with SAE activations $\mathbf{z}$, we clamp the activations of the top $n \triangleq \lfloor f \cdot N \rfloor$ TCAV-aligned features to their corresponding scalar values in the target centroid $\mathbf{c}_{\text{target}}$. This produces the intervened sparse activation $\mathbf{z}^{*}(f)$, which is then decoded back into embedding space:
    \begin{equation}
        \hat{\mathbf{a}}^*(f) = \mathbf{W}_{\text{dec}}\,\mathbf{z}^*(f) + \mathbf{b}_{\text{dec}}
    \label{eq:sae_decode}
    \end{equation}
\end{enumerate}

\subsection{Evaluating "steerability"}\label{sec:evaluating:steerability}
\label{sec:steering_eval}

\begin{enumerate} 
\setcounter{enumi}{3}
    \item \textbf{Frozen probe evaluation.} We score the decoded embeddings $\hat{\mathbf{a}}^*(f)$ using two linear probes (target and off-target). Crucially, these probes are fit once on clean ($f=0$) decoded embeddings and held frozen across the sweep; they act solely as measurement instruments to read out the result of the clamp, completely isolated from the substitution loop.
    
    \item \textbf{Excess Selectivity Scalar.} We quantify the quality of the intervention by the integrated area between the off-target and target performance curves:
    \begin{equation}
        \Delta(\text{rank}) = \int_0^1 \left( \mathrm{AUROC}_{\text{off-target}}(f) - \mathrm{AUROC}_{\text{target}}(f) \right) \mathrm{d}f
    \label{eq:selectivity}
    \end{equation}
    
    To account for random feature suppression which may degrade one probe faster than the other, we report the \textbf{excess selectivity} $\tilde{\Delta}$ as the difference between the TCAV-ranked area and the expected area under uniformly random feature permutations $\pi$:
    \begin{equation}
        \tilde{\Delta} = \Delta(\text{TCAV}) - \mathbb{E}_{\pi}[\Delta(\pi)]
    \label{eq:excess_selectivity}
    \end{equation}
\end{enumerate}

A value of $\tilde{\Delta} > 0$ indicates a selective push that successfully erases the target concept while sparing off-target axes. Conversely, a value of $\tilde{\Delta} \approx 0$ diagnoses a "wrecking-ball" regime: the intervention is no more selective than random noise, implying that the target concept is so entangled with the global embedding structure that it cannot be independently manipulated. This protocol operationalizes the selectivity criteria used in closed-form concept erasure~\cite{belrose2023leace} and amnesic probing~\cite{elazar2021amnesic} for the SAE latent space. Finally, decoding the intervened embedding $\hat{\mathbf{a}}^*(f)$ through the spectral decoder (Section~\ref{sec:stage-i}) yields domain-readable spectral signatures of the intervention (Figure~\ref{fig:spectrum-steering}).

\subsection{Choosing the operating point per model}
\label{sec:operating-point}

We identify an optimal layer $\ell^*$ and SAE expansion ratio $E^*$ for Stages II--IV by sweeping all transformer layers and $E \in \{1, 2, 4, 8, 16, 32, 64\}$. To match the active feature fraction across scales, sparsity scales linearly as $k = 8E$. We select the configuration that maximizes monosemanticity while penalizing dead features:
\begin{equation}
    (\ell^*, E^*) = \arg\max_{\ell, E} \left( \mathrm{separable}_{\ell,E} - \mathrm{dead}_{\ell,E} \right),
\label{eq:operating_point}
\end{equation}

constrained to layers where target concepts achieve model-level TCAV significance ($p < 0.05$). This taxonomy-driven metric efficiently isolates the model's most transparent regime (Figure~\ref{fig:operating-point}). While this identifies the optimal single coordinate, our computational capacity allows us to fix $E = E^*$ and explore various layers in subsequent interventional analyses to map the encoder's full representational trajectory.

\section{Results}

\subsection{Binary finetune performance and SAE faithfulness}

All three encoders separate the binary normal/abnormal target after end-to-end finetuning. We finetune each encoder under 5-fold subject-disjoint cross-validation and report the native classifier head's test performance, that is, the predictor we would actually deploy. As a faithfulness check on the SAE, we replace the encoder's layer-$\ell^*$ activations with the TopK-SAE reconstruction and re-evaluate the same head on the same test windows: classification AUROC is preserved within $0.017$ on all three encoders (Table~\ref{tab:layer-sweep-summary}), confirming that the SAE captures the task-relevant structure of the representation rather than discarding it.

\begin{table}[h]
  \centering
  \caption{\textbf{Layer-averaged SAE-faithfulness summary (5-fold CV).} All values are mean $\pm$ std across 5 subject-disjoint folds. Baseline metrics (Balanced Accuracy, F1-Score, and AUROC) are computed for the baseline (no-SAE) linear probe. Mean AUROC (SAE) is the per-layer test AUROC averaged across all transformer blocks of the encoder ($\pm$ is the mean per-fold std averaged across layers, so the unit matches the no-SAE column). Mean $\Delta$ is the average AUROC drop from baseline across all layers, with the std taken over layers; Max $\Delta$ is the worst-layer drop. The same 5-fold protocol underlies Figure~\ref{fig:layer-sweep}.}
  \label{tab:layer-sweep-summary}
  \footnotesize
  \setlength{\tabcolsep}{4pt}
  \resizebox{\textwidth}{!}{%
\begin{tabular}{lrcccccc}
    \toprule
    \textbf{Encoder} & \textbf{\# L} & \textbf{Balanced Acc.} & \textbf{F1-Score} & \textbf{AUROC (no SAE)} & \textbf{AUROC (SAE)} & \textbf{Mean $\Delta$} & \textbf{Max $\Delta$} \\
    \midrule
    SleepFM & 3  & $0.901 \pm 0.019$ & $0.866 \pm 0.047$ & $0.954 \pm 0.013$ & $0.951 \pm 0.015$ & $0.003 \pm 0.002$ & $0.005$ \\
    REVE    & 22 & $0.890 \pm 0.018$ & $0.857 \pm 0.044$ & $0.944 \pm 0.009$ & $0.944 \pm 0.009$ & $0.000 \pm 0.001$ & $0.006$ \\
    LaBraM  & 12 & $0.875 \pm 0.019$ & $0.835 \pm 0.045$ & $0.939 \pm 0.017$ & $0.933 \pm 0.020$ & $0.006 \pm 0.006$ & $0.017$ \\
    \bottomrule
\end{tabular}
  }
\end{table}

\begin{figure}[h]
  \centering
  \includegraphics[width=0.85\textwidth]{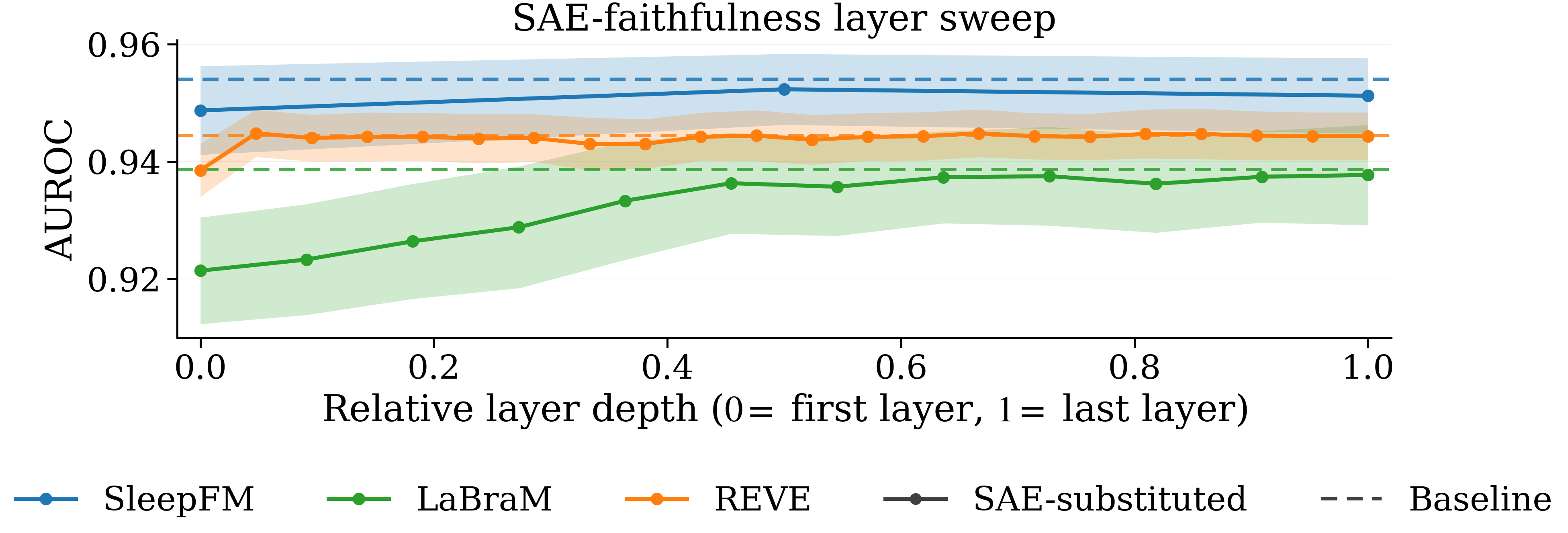}
  \caption{\textbf{SAE-faithfulness layer sweep.} Test AUROC of a linear probe trained via 5-fold cross-validation on mean-pooled embeddings of each finetuned encoder. During inference, layer-$\ell$ activations are replaced by their TopK-SAE reconstructions as $\ell$ sweeps through every transformer block. Shaded bands represent 95\% confidence intervals across the CV folds; the dotted horizontal lines indicate the no-SAE baseline mean. The SAE substitution consistently remains within the baseline performance envelope across all layers and architectures.}
  \label{fig:layer-sweep}
\end{figure}

The faithfulness check generalises beyond the operating layer. Figure~\ref{fig:layer-sweep} sweeps the SAE-substitution AUROC across every layer of every encoder: SleepFM stays within $0.005$ of its no-SAE baseline at all 3 layers, REVE within $0.006$ across all 22 layers, and LaBraM within $0.017$ across all 12 layers. Interestingly, we see an increase in performance over LaBraM's 12 layers, but from the fourth layer of LaBraM, and for all layers of SleepFM and REVE, the SAE substitution is statistically indistinguishable from the intact encoder regardless of where it is inserted; not just at $\ell^*$.

\begin{figure}[h]
  \centering
  \includegraphics[width=\linewidth]{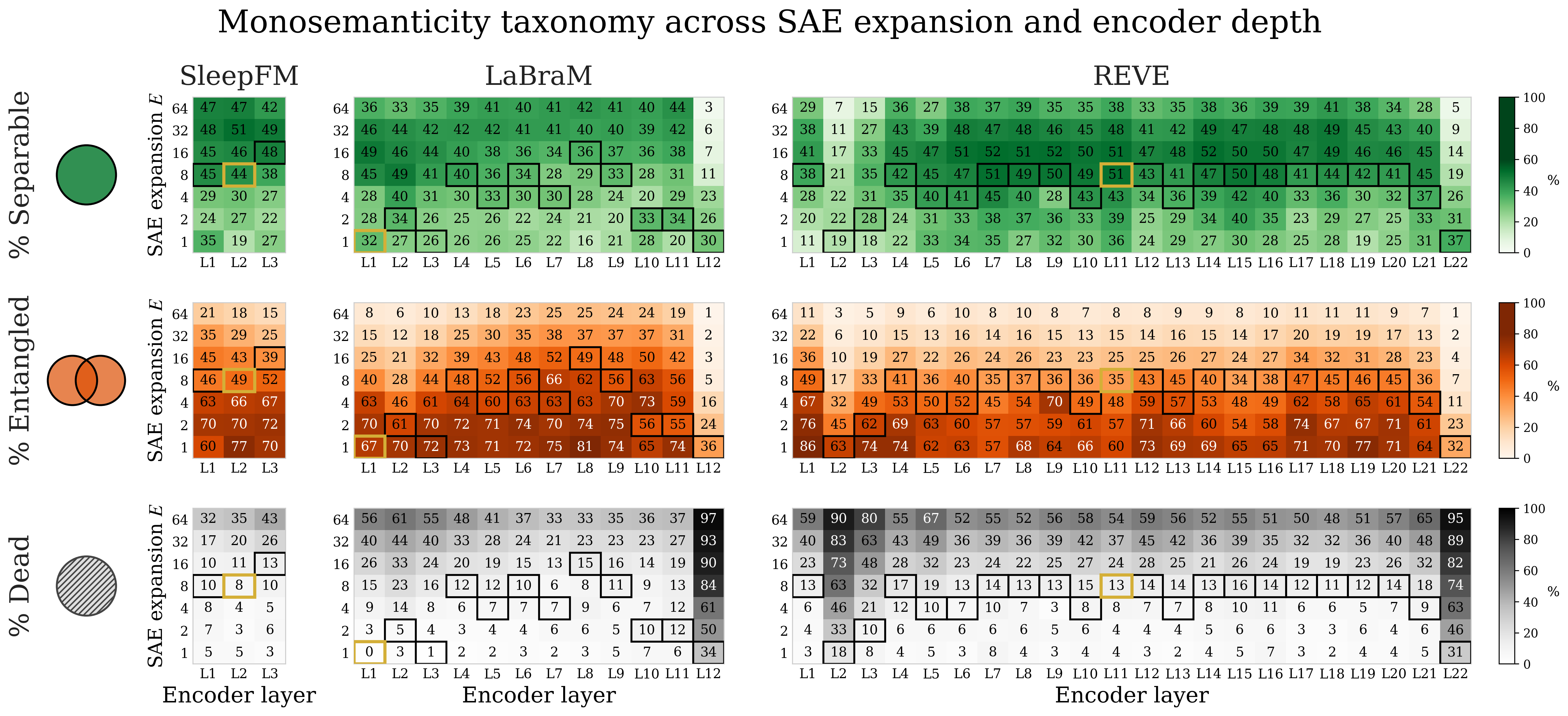}
  \caption{\textbf{Monosemanticity taxonomy across SAE expansion and encoder depth.} Each cell reports the fraction of concept-enriched SAE features in one of three taxonomy classes ({Separable}: monosemantic; {Entangled}: polysemantic co-activations; {Dead}: semantically uninformative/inactive). Columns represent encoders (SleepFM, LaBraM, REVE), with x-axes indexing the encoder layer and y-axes indexing expansion factor $E \in \{1, 2, 4, 8, 16, 32, 64\}$. For each layer, we highlight the expansion factor that maximizes the difference between separable and dead feature fractions. The optimum $(\ell^*, E^*)$ is golden. Find an overview of how $E$ translates to dictionary size in Figure \ref{fig:sae-dictionary-size}.}
  \label{fig:operating-point}
\end{figure}

\begin{figure}[h]
  \centering
  \includegraphics[width=\linewidth]{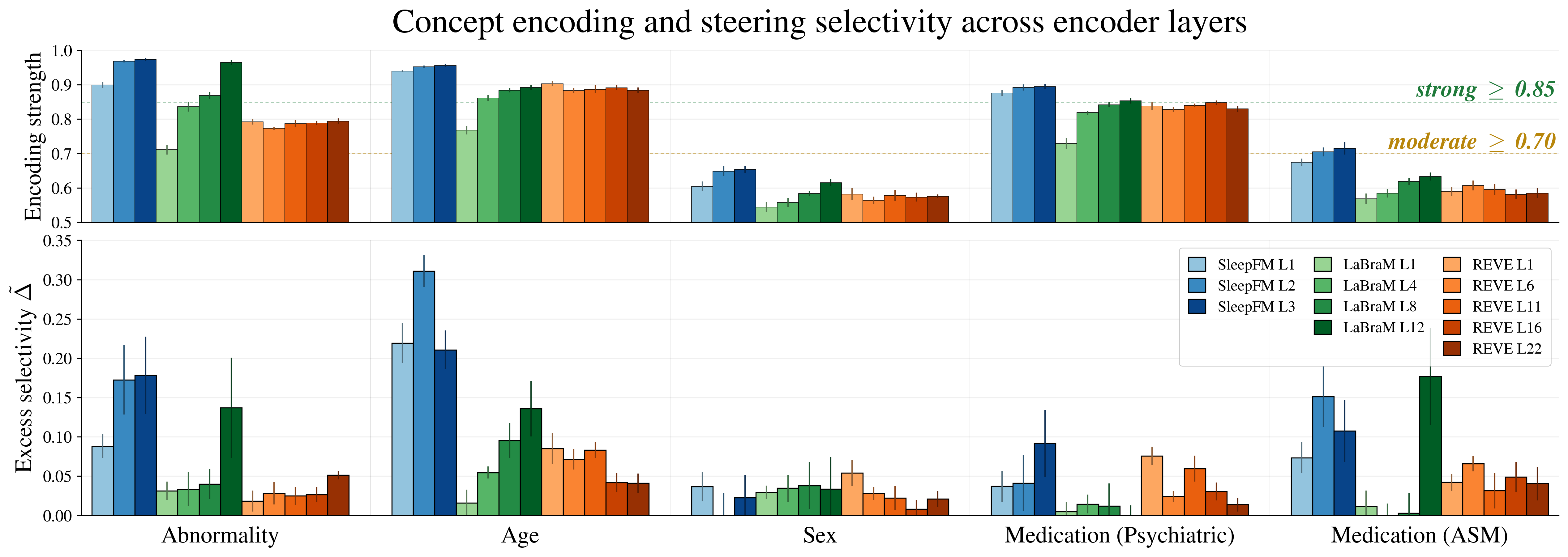}
\caption{\textbf{Concept encoding strength and steering selectivity.}
\textbf{Top:} Encoding strength ($\mathrm{AUROC}_0$) measured via per-layer linear probes fit to the clean SAE-decoded reconstructions.
\textbf{Bottom:} Excess selectivity ($\tilde{\Delta}$), quantifying the integrated asymmetry between target and off-target probe degradation under TCAV-ranked clamping (Section~\ref{sec:steering}). For \textit{abnormality} as target, we use \textit{age} as off-target. For all other targets, we use \textit{abnormality} as off-target. Together, these metrics map the representational landscape of each model, distinguishing selectively steerable features from highly entangled ones. Exactly how excess selectivity $\tilde{\Delta}$ is computed is illustrated in Figure \ref{fig:steerability-examples} through examples.}
  \label{fig:steerability}
\end{figure}

\subsection{Spectral decoder reconstruction.} The spectral decoder reliably recovers token-level amplitude spectra from frozen embeddings across all architectures. Overall amplitude $R^2$ reaches $0.816 \pm 0.004$ for SleepFM, $0.783 \pm 0.004$ for REVE, and $0.772 \pm 0.004$ for LaBraM. 
Conversely, phase remains unrecoverable across all models (cosine similarity $\le 0.22$), directly reflecting the time-translation invariance inherent to their self-supervised pretraining objectives. All margins represent 95\% bootstrap intervals.


\subsection{Operating points transfer across encoders}

Figure~\ref{fig:operating-point} shows the per-encoder operating points                                                  
  $(\ell^*, E^*)$ selected by the procedure of                
  Section~\ref{sec:operating-point}. Despite the three encoders differing in depth (3/12/22 transformer blocks), embedding dimension 
  (128/200/512) and training objective, the recipe converges on a common modal expansion across each network: $E^* = 8$ wins at 15/22 REVE layers, 3/12 LaBraM layers, and 2/3 SleepFM layers. High separation at low $E$ concentrates at the boundaries: the encoders' first and last transformer blocks collapse to a smaller dictionary (LaBraM L1, L3, L12 and REVE L2, L22 at $E^* = 1$; LaBraM L2, L10, L11 and REVE L2 at $E^* = 2$), consistent with the classification block compressing representations into a low-dimensional task-aligned subspace. SleepFM, with only three transformer blocks, does not compress the representations.



\begin{figure}[h]
  \centering
  \includegraphics[width=\linewidth]{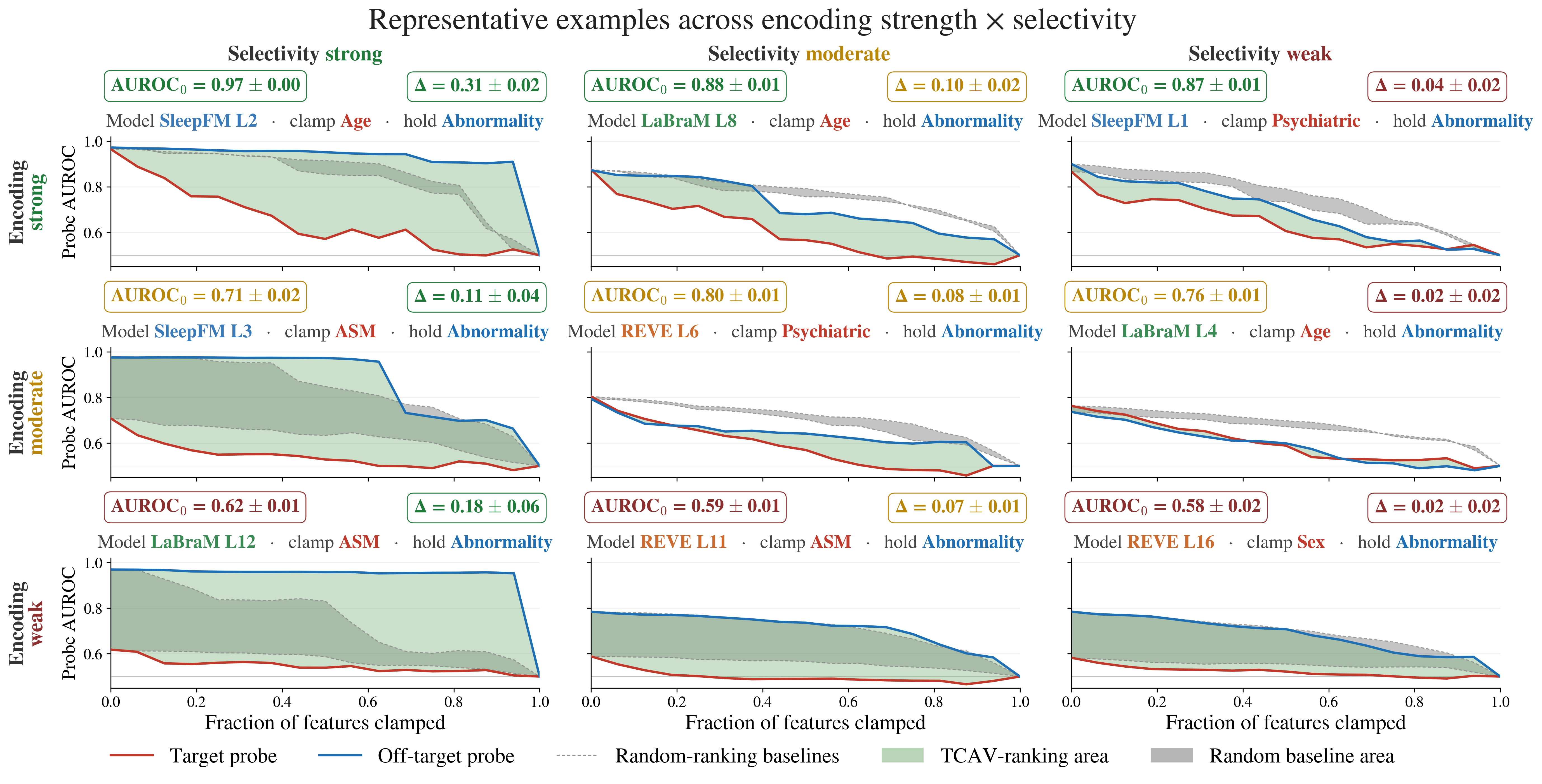}
  \caption{\textbf{Steering sweeps across the encoding--selectivity landscape.} Nine representative configurations from Figure~\ref{fig:steerability} arranged by encoding strength (rows) and selectivity $\tilde{\Delta}$ (columns). Panels track target (red) and off-target (blue: \textit{abnormality}) AUROC as clamping fraction $f$ increases. Green shading highlights the TCAV-driven selectivity gain while grey shading highlights the random-clamping baseline. The grid contrasts canonical selective steering (top-left: target collapses, off-target is spared) with the ``wrecking-ball'' failure mode (top two cells in the right column: both concepts degrade through entanglement, $\tilde{\Delta} \approx 0$). Encoder colors match Figure~\ref{fig:steerability}.}
  \label{fig:steerability-examples}
\end{figure}

\subsection{Cross-model concept selectivity}

We apply the substitution-sweep protocol of Section~\ref{sec:steering} to every (encoder, layer, concept) triplet and report excess selectivity $\tilde{\Delta}$ alongside the clean-baseline encoding strength (Figure~\ref{fig:steerability}). The two rows together discriminate three regimes:

\begin{itemize}
    \item \textbf{Encoded and steerable:} High AUROC$_0$ + high $\tilde{\Delta}$.
    \item \textbf{Encoded but not steerable:} High AUROC$_0$ + $\tilde{\Delta} \approx 0$.
    \item \textbf{Weakly encoded:} AUROC$_0 < 0.7$.
\end{itemize}

To better illustrate the underlying probe-AUROC sweeps that compute excess selectivity ($\tilde{\Delta}$), in Figure~\ref{fig:steerability-examples}, we unpack nine representative cells from the cross-model summary in Figure~\ref{fig:steerability}. To understand this grid, the axes should be read as two distinct constraints on the intervention:

\textbf{Rows define the interventional headroom.} The baseline encoding strength dictates the maximum possible impact of the clamp. If a concept is weakly encoded, there is no signal to erase; the target probe is already near random guessing, rendering the sweep flat and uninformative.

\textbf{Columns map the selectivity regimes.} This axis illustrates the continuum from surgical concept removal to unselective collapse. We categorize this into three distinct behaviors:
 
\begin{itemize}
    \item \textbf{Strong selectivity:} The target probe (red) degrades rapidly under TCAV-ranked clamping while the off-target probe (blue) remains robust. Crucially, the target degrades much faster than the random-clamping baseline, yielding a large selectivity area (green) while the random-area expectation (grey) is insignificant.
    \item \textbf{Moderate selectivity:} TCAV ranking isolates the concept faster than random noise, but the off-target probe is dragged down alongside it, indicating partial entanglement.
    \item \textbf{Weak selectivity:} The green area shrinks to or below the grey area. Target and off-target representations degrade in lockstep. The intervention removes signal, but it acts as a global ``wrecking ball'' rather than a targeted push.
\end{itemize}

Ultimately, this grid provides the qualitative, mechanistic proof for the scalar $\tilde{\Delta}$ summaries in Figure~\ref{fig:steerability}. It allows the reader to visually verify that a high $\tilde{\Delta}$ truly guarantees an asymmetric, concept-aligned steering effect, rather than a uniform collapse of the layer's representation.

 Looking at Figures~\ref{fig:steerability} and ~\ref{fig:steerability-examples}, examples of encoded and steerable concepts are \textit{abnormality} and \textit{age} in SleepFM L2/L3 and \textit{ASM} in LaBraM L12. An example of an encoded but non steerable concept is \textit{abnormality} across all representative REVE layers. Finally, an example of a weakly encoded concept is sex across all encoders; which functions as the implicit negative control. Two "wrecking-ball" examples can be found in Figure~\ref{fig:steerability-examples} in the two top cells in the right column (\textit{psychiatric medicine} in SleepFM L1 and \textit{age} in LaBraM L4): both concepts degrade through entanglement.

\subsection{A worked spectrum-level example}

\begin{figure}[h]
  \centering
  \includegraphics[width=\linewidth]{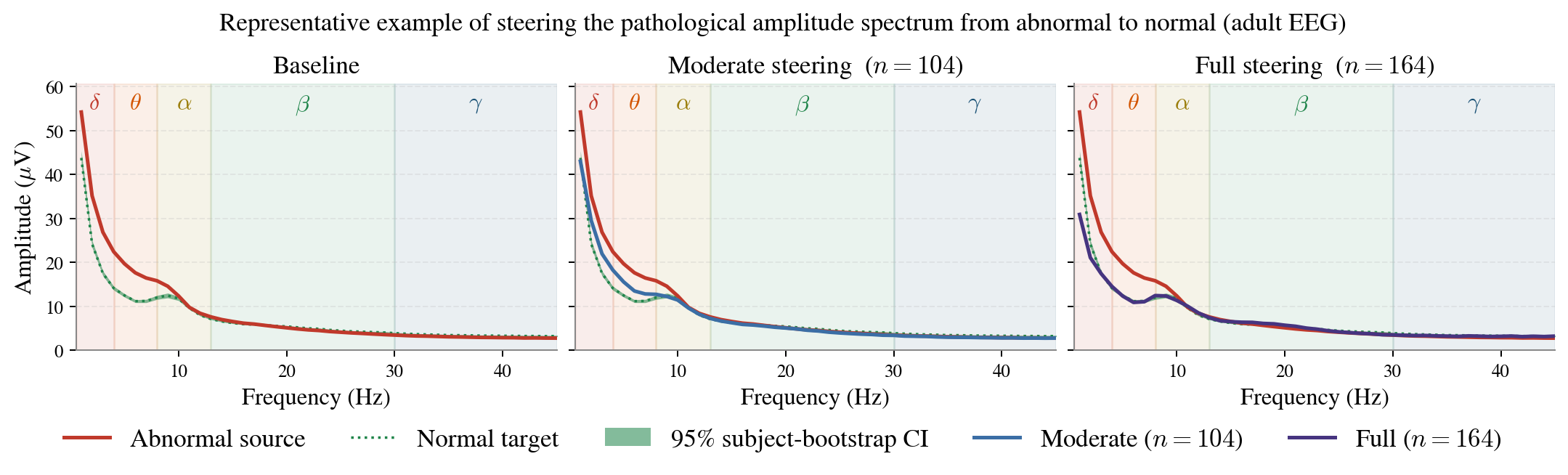}
  \caption{\textbf{Spectrum-level concept steering (abnormal $\to$ normal).} SleepFM layer 2 ($E=8$). Shading denotes 95\% bootstrap CIs on the target mean. \textbf{Left:} Baseline abnormal source vs.\ normal target centroid. \textbf{Centre \& Right:} The decoded spectrum after clamping the top $n = 104$ and $n = 164$ TCAV-aligned features, respectively, to the target centroid.}
  \label{fig:spectrum-steering}
\end{figure}

Figure~\ref{fig:spectrum-steering} grounds the abstract selectivity metrics (Figure~\ref{fig:steerability}) in domain-readable signatures. Steering a representative abnormal adult EEG toward a normal target corrects canonical pathological markers: clamping the top TCAV-aligned features actively collapses the elevated pathological $\delta$ and $\theta$ power while restoring the suppressed $\alpha$ peak. By $n = 164$, the intervened spectrum fully merges into the healthy reference band across the entire 0.5--45~Hz range, providing clinical experts with physical proof of targeted concept manipulation.

\section{Discussion and limitations}

\textbf{Decoder limitations.} Our spectral decoder recovers amplitude but cannot reconstruct phase, as current pretraining objectives discard precise temporal morphology in favor of time-translation invariance. Consequently, while we can steer broad frequency distributions, we cannot generate exact time-domain waveforms (e.g., specific spike-wave morphologies). High-fidelity time-domain steering will ultimately require phase-aware foundation models or significantly more expressive decoders.

\textbf{Separability vs.\ selectivity.} Using observational monosemanticity, we assume that separable features provide clean interventional handles. However, separability does not guarantee selectivity: clamping an observationally pure feature can still trigger a collapse. Future work should evaluate whether optimizing directly for interventional metrics ($\tilde{\Delta}$) yields safer operating points.

\textbf{Concept taxonomy is investigator-defined.} 
Because our steering protocol (Section~\ref{sec:steering}) requires pre-specified concepts, our evaluation is strictly bounded by available cohort metadata. The pipeline verifies rather than discovers. Although inherent to all probe-based methods, this restricts how comprehensively a label-poor dataset can audit a model's latent space.

\newpage
\section{Conclusion}
We propose a model-agnostic interpretability framework for auditing the representational health of EEG foundation models. By integrating Sparse Autoencoders, concept steering, and spectral decoding across three architecturally distinct encoders, the pipeline maps the semantic taxonomy of transformer blocks; isolating where clinical concepts consolidate into selectively steerable features, and where persistent entanglement triggers ``wrecking-ball'' failures during intervention. By decoding latent representations back into domain-readable spectral signatures, such as focal slowing or alpha-band suppression, we bridge the gap between abstract embedding spaces and clinical physiology, grounding model attributions in mechanistic, frequency-level explanations.

\paragraph{Broader Impact.} This framework is intended as a diagnostic tool to increase transparency and clinical trust in EEG foundation models. By surfacing representational failures such as age–pathology entanglement, it can help practitioners identify and mitigate latent demographic biases. As with all medical AI, frameworks like this should be rigorously validated.



\newpage
\bibliographystyle{unsrtnat}
\bibliography{refs}

\newpage
\appendix

\section{Technical appendices and supplementary material}

\begin{table}[h]
\centering
\small
\caption{Notation reference.}
\begin{tabular}{lllr}
\toprule
\textbf{Symbol} & \textbf{Type / shape} & \textbf{Definition} & \textbf{First used} \\
\midrule
\multicolumn{4}{l}{\textit{Encoder}} \\
$d$ & scalar $\in \mathbb{Z}^+$ & Embedding dimension (128 / 200 / 512 for SleepFM / LaBraM / REVE) & Eq.~\ref{eq:sae} \\
$\mathbf{a}$ & vector $\in \mathbb{R}^d$ & Layer activation extracted from the frozen encoder & Eq.~\ref{eq:sae} \\
$\mathbf{t}$ & vector $\in \mathbb{R}^d$ & Token embedding passed to the spectral decoder & Eq.~\ref{eq:sd} \\
$\ell$ & scalar $\in \mathbb{Z}^+$ & Transformer layer index & Sec. \ref{sec:stage-ii} \\
$\ell^*$ & scalar $\in \mathbb{Z}^+$ & Optimal layer selected by Eq.~\ref{eq:operating_point} & Eq.~\ref{eq:operating_point} \\
\midrule
\multicolumn{4}{l}{\textit{SAE architecture}} \\
$E$ & scalar $\in \mathbb{Z}^+$ & Expansion ratio & Sec. \ref{sec:stage-ii} \\
$N \triangleq d \cdot E$ & scalar $\in \mathbb{Z}^+$ & Dictionary size (number of SAE features) & Eq.~\ref{eq:sae} \\
$k$ & scalar $\in \mathbb{Z}^+$ & TopK sparsity budget; $k = k_0 \cdot E$, $k_0 = 8$ & Eq.~\ref{eq:sae} \\
$k_0$ & scalar $\in \mathbb{Z}^+$ & Base sparsity constant; $k_0=8$, so $k/N = k_0/d$ is constant & Sec. \ref{sec:stage-ii} \\
$W_{\text{enc}}$ & matrix $\in \mathbb{R}^{N \times d}$ & SAE encoder weight matrix & Eq.~\ref{eq:sae} \\
$W_{\text{dec}}$ & matrix $\in \mathbb{R}^{d \times N}$ & SAE decoder; columns $\mathbf{w}_i \in \mathbb{R}^d$ are unit-norm directions & Eq.~\ref{eq:sae} \\
$\mathbf{w}_i$ & vector $\in \mathbb{R}^d$ & $i$-th column of $W_{\text{dec}}$; unit-norm decoder direction & Eq.~\ref{eq:sae} \\
$b_{\text{dec}}$ & vector $\in \mathbb{R}^d$ & SAE decoder bias & Eq.~\ref{eq:sae} \\
$\mu_\ell,\,\sigma_\ell$ & vectors $\in \mathbb{R}^d$ & Per-dimension mean and std used to normalise activations & Eq.~\ref{eq:sae} \\
$\mathbf{z}$ & vector $\in \mathbb{R}^N$ & Sparse SAE feature activation (exactly $k$ non-zero entries) & Eq.~\ref{eq:sae} \\
$\hat{\mathbf{a}}$ & vector $\in \mathbb{R}^d$ & SAE reconstruction of the encoder activation & Eq.~\ref{eq:sae} \\
$E^*$ & scalar $\in \mathbb{Z}^+$ & Optimal expansion ratio selected by Eq.~\ref{eq:operating_point} & Eq.~\ref{eq:operating_point} \\
\midrule
\multicolumn{4}{l}{\textit{Spectral decoder}} \\
$\mathrm{SD}$ & $\mathbb{R}^d \!\to\! \mathbb{R}^F \!\times\! \mathbb{R}^{2F}$ & Shallow MLP mapping a token embedding to amplitude and phase & Eq.~\ref{eq:sd} \\
$F$ & scalar $\in \mathbb{Z}^+$ & Number of frequency bins ($F = 64$) & Eq.~\ref{eq:sd} \\
$\hat{A}_\nu$ & scalar $\in \mathbb{R}$ & Predicted amplitude at frequency bin $\nu$ & Eq.~\ref{eq:sd} \\
$\hat{\varphi}_\nu$ & scalar $\in [0,2\pi)$ & Phase at bin $\nu$; parameterised as $(\cos\hat{\varphi}_\nu, \sin\hat{\varphi}_\nu)$ & Eq.~\ref{eq:sd} \\
\midrule
\multicolumn{4}{l}{\textit{Concept attribution (TCAV)}} \\
$\mathcal{C}$ & label & Clinical concept (abnormality, age group, sex, medication) & Sec. \ref{sec:stage-iii} \\
$\mathbf{v}_C$ & vector $\in \mathbb{R}^d$ & Concept Activation Vector; unit-norm direction in activation space & Sec. \ref{sec:background_tcav} \\
$S_C(x)$ & scalar $\in \mathbb{R}$ & TCAV sensitivity score for example $x$ under concept $C$ & Eq.~\ref{eq:tcav} \\
$\mathrm{TCAV}_C$ & scalar $\in [0,1]$ & Fraction of concept examples with $S_C(x)>0$; above 0.5 $\Rightarrow$ encoded & Eq.~\ref{eq:tcav} \\
$N_{\text{rand}}$ & scalar $\in \mathbb{Z}^+$ & Number of random-label null CAVs for significance testing; $N_{\text{rand}}=50$ & Sec. \ref{sec:background_tcav}  \\
$K_{fold}$ & scalar $\in \mathbb{Z}^+$ & Cross-validation folds for CAV training; $K_{fold}=10$ & Sec. \ref{sec:stage-iii} \\
\midrule
\multicolumn{4}{l}{\textit{Concept steering}} \\
$\mathrm{rank}_C(i)$ & function $\mathbb{Z}^+\!\to\!\mathbb{R}$ & CAV-alignment score for feature $i$: $|\mathbf{v}_C \cdot \mathbf{w}_i|$ & Eq.~\ref{eq:rank_features} \\
$X_{\text{target}}$ & dataset & Target pool for the steering intervention & Eq.~\ref{eq:c_target} \\
$\mathbf{c}_{\text{target}} \in \mathbb{R}^N$ & vector $\in \mathbb{R}^N$ & Target-concept centroid; mean SAE activations over $X_{\text{target}}$ & Eq.~\ref{eq:c_target} \\
$f$ & scalar $\in [0,1]$ & Clamping fraction & Eq.~\ref{eq:sae_decode} \\
$n \triangleq \lfloor f \cdot N \rfloor$ & scalar $\in \mathbb{Z}^+$ & Number of features clamped (also used bare in Section 4.5 / Fig.~6) & Eq.~\ref{eq:sae_decode} \\
$\mathbf{z}^*(f)$ & vector $\in \mathbb{R}^N$ & Intervened activation after clamping top-$n$ features & Eq.~\ref{eq:sae_decode} \\
$\hat{\mathbf{a}}^*(f)$ & vector $\in \mathbb{R}^d$ & SAE-decoded embedding of $\mathbf{z}^*(f)$ & Eq.~\ref{eq:sae_decode} \\
$\Delta(\text{rank})$ & scalar $\in \mathbb{R}$ & Raw selectivity area & Eq.~\ref{eq:selectivity} \\
$\tilde{\Delta}$ & scalar $\in \mathbb{R}$ & Excess selectivity $= \Delta(\text{TCAV}) - \mathbb{E}_\pi[\Delta(\pi)]$ & Eq.~\ref{eq:excess_selectivity} \\
\midrule
\multicolumn{4}{l}{\textit{Hyperparameter selection}} \\
$\text{separable}_{\ell,E}$ & scalar $\in [0,1]$ & Fraction of features classified as separable at layer $\ell$, expansion $E$ & Eq.~\ref{eq:operating_point} \\
$\text{dead}_{\ell,E}$ & scalar $\in [0,1]$ & Fraction of features classified as dead at layer $\ell$, expansion $E$ & Eq.~\ref{eq:operating_point} \\
\bottomrule
\end{tabular}
\end{table}

\begin{table}[h]
\centering
\scriptsize
\caption{Encoder architectures and binary fine-tuning hyperparameters.
All three encoders are finetuned end-to-end on the binary
(normal vs.\ abnormal) label using AdamW with two-phase training:
a head-only warm-up phase followed by full unfreezing.
Learning rate is dropped when the encoder is unfrozen and the weight decay
is increased.}
\label{tab:encoder_hparams}
\setlength{\tabcolsep}{4pt}
\begin{tabular}{l p{2.8cm} p{3.2cm} p{2.8cm}}
\toprule
 & \href{https://www.nature.com/articles/s41591-025-04133-4}{SleepFM} & \href{https://arxiv.org/abs/2510.21585}{REVE-Base} & \href{https://arxiv.org/abs/2405.18765}{LaBraM-Base} \\
 \midrule
 \emph{Where can I find the encoder?}
  & \href{https://github.com/zou-group/sleepfm-clinical}{Github}
  & \href{https://huggingface.co/brain-bzh/reve-base}{Hugging Face}
  &  \href{https://braindecode.org/dev/generated/braindecode.models.Labram.html}{braindecode} and \href{https://github.com/935963004/LaBraM}{Github} \\
\midrule
\multicolumn{4}{l}{\emph{Architecture}} \\
Backbone
  & {SetTransformer}
  & {ViT}
  & {NeuralTransformer} \\
Embed.\ dim         & 128                & 512                & 200 \\
 Transformer layers      & 3                    & 22                                   & 12 \\                                                        
  Attention heads         & 8                    & 8                                    & 10 \\                                                        
  Head dim                & 16                   & 64                                   & 20 \\                                                        
  MLP ratio               & 16       & 2.66 (GeGLU)                         & 4.0 \\                                                       
  Patch size (samples)    & 128                  & 200                                  & 200 \\ 
Tokens per window       & 60                   & $19 \times 66$ (overlapping) & $19 \times 60$ \\
  Token receptive field   & 1.00\,s              & 1.00\,s                                    & 1.00\,s \\                                             
  Token stride            & 1.00\,s              & 0.90\,s                                    & 1.00\,s \\
\midrule
\multicolumn{4}{l}{\emph{Input data}} \\
Dataset             & Binary (normal/abnormal)           & Binary (normal/abnormal)           & Binary (normal/abnormal) \\
Sample rate (Hz)    & 128                & 200                & 200 \\
Window length   & 60                 & 60                 & 60 \\
Channels            & 27 (10-20)   & 27 (10-20)         & 19 (10-20) \\
\midrule
\multicolumn{4}{l}{\emph{Finetuning}} \\
 Pooling          & temporal\_pooling        & mean over $C\!\cdot\!S$ tokens     & mean over patch tokens \\                       
  Head             & \texttt{Linear(128,1)}            & \texttt{Linear(512,1)}             & \texttt{Linear(200,1)} \\                                
  Loss             & BCE & BCE & BCE \\
  \midrule
\multicolumn{4}{l}{\emph{Optimisation}} \\
Optimiser           & AdamW              & AdamW              & AdamW \\
Total epochs        & 12                 & 10                 & 15 \\
Head-only w/u ep.   & 2                  & 2                  & 3 \\
Batch size          & 32                 & 8                  & 8 \\
LR (head phase)     & $10^{-3}$          & $10^{-3}$          & $10^{-3}$ \\
LR (full phase)     & $10^{-4}$          & $10^{-4}$          & $10^{-4}$ \\
WD (head/full)      & 0.01\,/\,0.05      & 0.01\,/\,0.05      & 0.01\,/\,0.05 \\
Grad-norm clip      & 1.0                & 1.0                & 1.0 \\
\bottomrule
\end{tabular}
\end{table}

\begin{figure}[htbp]
  \centering
  \includegraphics[width=\linewidth]{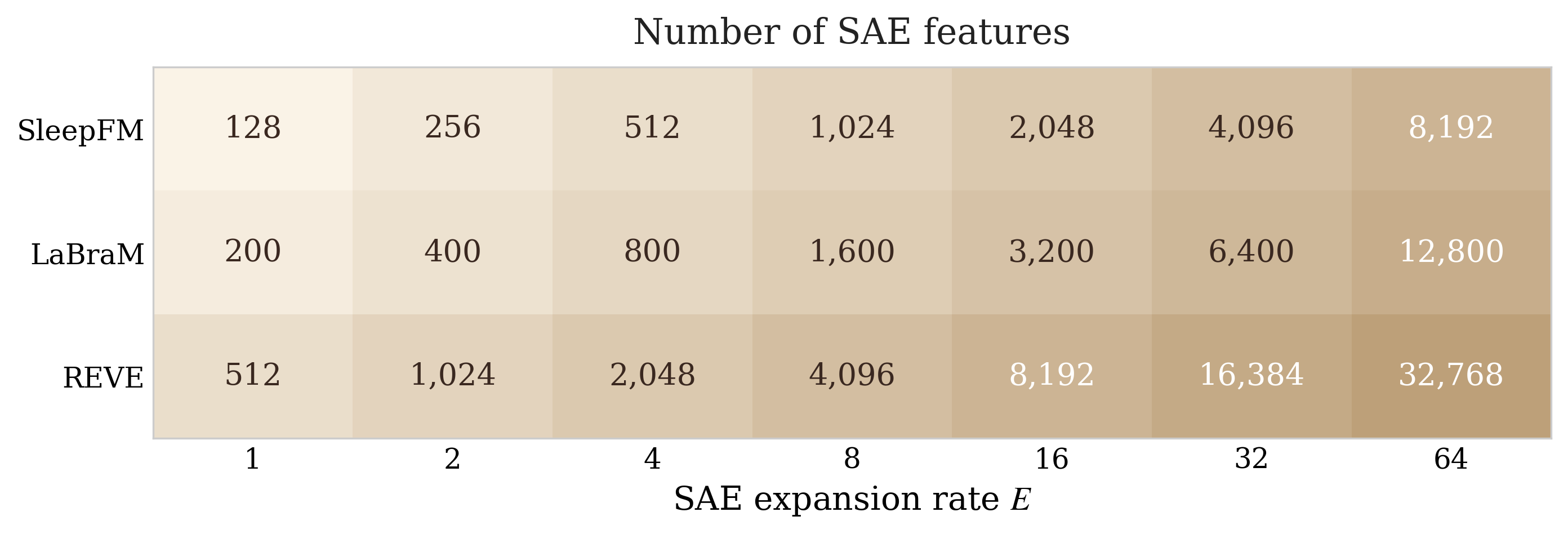}
  \caption{\textbf{SAE dictionary size across encoders and expansion rates.}
  Each cell gives the number of learned SAE features, which equals the encoder's
  embedding dimension ($d_{\text{enc}} = 128$ for SleepFM, $200$ for LaBraM,
  $512$ for REVE) times the expansion rate $E$. Because REVE is $4\times$ wider
  than SleepFM, an $E{=}1$ REVE SAE already exceeds the size of an $E{=}4$
  SleepFM SAE, and the $E{=}64$ REVE configuration spans $32{,}768$ features.
  This asymmetry should be kept in mind when comparing taxonomy fractions
  (Fig.~\ref{fig:operating-point}) across encoders at matched $E$: equal $E$ does
  \emph{not} mean equal capacity.}
  \label{fig:sae-dictionary-size}
\end{figure}

\end{document}